\documentclass[10pt,twocolumn,letterpaper]{article}

\usepackage{iccv}
\usepackage{times}
\usepackage{epsfig}
\usepackage{graphicx}
\usepackage{amsmath}
\usepackage{amssymb}
\usepackage{multirow}
\usepackage{tabularx}
\usepackage{booktabs,fixltx2e}
\usepackage{threeparttablex}
\usepackage{adjustbox}
\usepackage[labelfont=bf]{caption}
\usepackage{siunitx}
\usepackage[fleqn]{mathtools}
\usepackage{float}
\usepackage{cite}
\usepackage[ruled,vlined,linesnumbered]{algorithm2e}
\usepackage{algpseudocode}
\usepackage{adjustbox}

\def\figvspace{{\vspace{-4mm}}}

\newcommand{\Section}[1]{\vspace{-2mm} \section{#1} \vspace{-1mm}}
\newcommand{\SubSection}[1]{\vspace{-1mm} \subsection{#1} \vspace{-1mm}}

\newcommand{\Paragraph}[1]{\vspace{-0mm} \noindent \textbf{#1} \hspace{-0mm}}
\newcommand\T{{\hspace{-1pt}\intercal}}
\DeclareMathOperator*{\argmin}{arg\,min}

\usepackage[pagebackref=true,breaklinks=true,letterpaper=true,colorlinks,bookmarks=false]{hyperref}

 \iccvfinalcopy 

\def\iccvPaperID{793} 

\ificcvfinal\fi
\begin{document}

\title{Pose-Invariant 3D Face Alignment}

\author{Amin Jourabloo, Xiaoming Liu \\
Department of Computer Science and Engineering \\
Michigan State University, East Lansing MI 48824\\
{\{jourablo, liuxm\}@msu.edu}
}

\maketitle

\begin{abstract}
Face alignment aims to estimate the locations of a set of landmarks for a given image.
This problem has received much attention as evidenced by the recent advancement in both the methodology and performance.
However, most of the existing works neither explicitly handle face images with arbitrary poses, nor perform large-scale experiments on non-frontal and profile face images.
In order to address these limitations, this paper proposes a novel face alignment algorithm that estimates both $2$D and $3$D landmarks and their $2$D visibilities for a face image with an arbitrary pose.
By integrating a $3$D deformable model, a cascaded coupled-regressor approach is designed to estimate both the camera projection matrix and the $3$D landmarks. 
Furthermore, the $3$D model also allows us to automatically estimate the $2$D landmark visibilities via surface normals.
We gather a substantially larger collection of all-pose face images to evaluate our algorithm and demonstrate superior performances than the state-of-the-art methods.
\end{abstract}



\Section{Introduction}

This paper aims to advance {\it face alignment} in aligning face images with arbitrary {\it poses}.
Face alignment is a process of applying a supervised learned model to a face image and estimating the location of a set of facial landmarks, such as eye corners, mouth corners, etc~\cite{Cootes1994}.
Face alignment is a key module in the pipeline of most facial analysis algorithms, normally {\it after} face detection and {\it before} subsequent feature extraction and classification. 
Therefore, it is an enabling capability with a multitude of applications, such as face recognition~\cite{wagner2012toward}, expression recognition~\cite{bettadapura2012face}, etc. 

Given the importance of this problem, face alignment has been studied extensively since Dr.~Cootes' Active Shape Model (ASM) in the early $1990$s~\cite{Cootes1994}.
Especially in recent years, face alignment has become one of the most published subjects in top vision conferences~\cite{asthana2013robust, tzimiropoulos2014gauss,xiong2013supervised,xing2014towards,yang2013sieving,zhang2014coarse,luo2012hierarchical}.
The existing approaches can be categorized into three types: Constrained Local Model (CLM)-based approach (e.g.,~\cite{Cootes1994,Saragih2009a}), Active Appearance Model (AAM)-based approach (e.g., ~\cite{Matthews2004,Liu2008b}) and regression-based approach (e.g.,~\cite{valstar2010facial,cao2014face}), and an excellent survey of face alignment can be found in~\cite{wang2014facial}.

\begin{figure}[t!]
	\centering
	\includegraphics[width=1.0\linewidth]{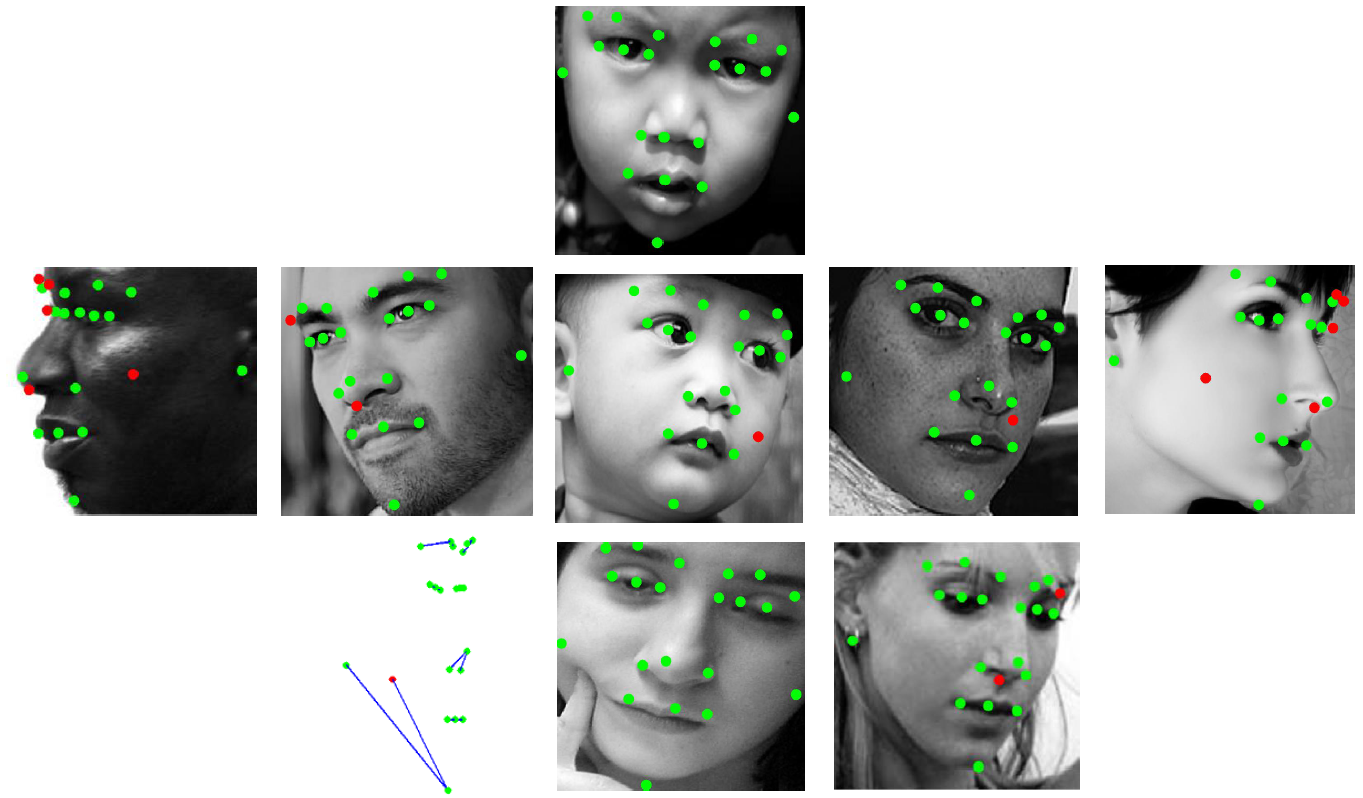}
	\caption{\small Given a face image with an arbitrary {\it pose}, our proposed algorithm automatically estimate the {\it $2$D location} and {\it visibilities} of facial landmarks, as well as {\it $3$D landmarks}.  The displayed $3$D landmarks are estimated for the image in the center. Green/red points indicate visible/invisible landmarks. } 
	\figvspace
	\label{fig:teaser}
\end{figure}

Despite the continuous improvement on the alignment accuracy, face alignment is still a very challenging problem, due to the non-frontal face {\it pose}, low image {\it quality}, {\it occlusion}, etc.
Among all the challenges, we identify the {\it pose invariant face alignment} as the one deserving substantial research efforts, for a number of reasons.
First, face detection has substantially advanced its capability in detecting faces in all poses, including profiles~\cite{zhang2010survey}, which calls for the subsequent face alignment to handle faces with arbitrary poses.
Second, many facial analysis tasks would benefit from the robust alignment of faces at all poses, such as expression recognition and $3$D face reconstruction~\cite{Roth2015}.
Third, there are very few existing approaches that can align a face with any view angle, or have conducted extensive evaluations on face images across $\pm90^{\circ}$ yaw angles~\cite{yu2013pose,zhu2012face}, which is a clear {\it contrast} with the vast face alignment literature~\cite{wang2014facial}.

\begin{table*}[t!]

	\centering
	\caption{The comparison of face alignment algorithms in pose handling (Estimation error may has different definition.).}
	\resizebox{\textwidth}{!} 
{
	\begin{tabular}{l|c|c|c|c|c|c|c|c}
		\hline

	\multirow{2}{*}{Method} & $3$D & \multirow{2}{*}{Visibility}& \multirow{2}{*}{Pose-related database} & Pose & Training & Testing& Landmark & Estimation \\ 
& landmark& & & range & face \# & face \# &\# & error \\ \hline
		
RCPR~\cite{burgos2013robust} & No& Yes & COFW & frontal w.~occlu.  & $1,345$ & $507$ & $19$ & $8.5$ \\ \hline
CoR~\cite{yu2014consensus}& No& Yes & COFW; LFPW-O; Helen-O & frontal w.~occlu. & $1,345; 468; 402$ & $507;112; 290$  & $19;49;49$ & $8.5$ \\ \hline
TSPM~\cite{zhu2012face}& No & No & AFW & all poses& $2,118$ & $468$  & $6$& $11.1$  \\ \hline
CDM~\cite{yu2013pose}&No&No& AFW& all poses& $1,300$ & $468$  & $6$& $9.1$  \\ \hline
OSRD~\cite{xing2014towards}&No&No& MVFW& $<\pm40^{\circ}$ & $2,050$ & $450$  & $68$& N/A \\ \hline 
TCDCN~\cite{zhang2014facial}&No&No& AFLW, AFW & $<\pm60^{\circ}$ &$10,000$ & $3,000;\sim$$313$  & $5$ & $8.0; 8.2$  \\ \hline

PIFA& Yes& Yes & AFLW, AFW & all poses& $3,901$  & $1,299; 468$ & $21,6$ & $6.5; 8.6$  \\ \hline
	       \hline
	\end{tabular}
	} 
\label{tab:priorwork}
\end{table*}

Motivated by the needs to address the pose variation, and the lack of prior work in handling poses, as shown in Fig.~\ref{fig:teaser}, this paper proposes a novel regression-based approach for {\it pose-invariant face alignment}, which aims to estimate the {\it $2$D and $3$D location} of face landmarks, as well as their {\it visibilities} in the $2$D image, for a face image with {\it arbitrary pose} (e.g., $\pm90^{\circ}$ yaw).
By extending the popular cascaded regressor for $2$D landmark estimation, we learn two regressors for each cascade layer, one for predicting the update for the camera projection matrix, and the other for predicting the update for the $3$D shape parameter.
The learning of two regressors is conducted alternatively with the goal of minimizing the difference between the ground truth updates and the predicted updates. 
By assuming the $3$D surface normal of $3$D landmarks, we can automatically estimate the visibilities of their $2$D projected landmarks by inspecting whether the transformed surface normal has a positive $z$ value, and these visibilities are dynamically incorporated into the regressor learning such that only the local appearance of visible landmarks contribute to the learning.
Finally, extensive experiments are conducted on a large subset of AFLW dataset~\cite{koestinger11b} with a wide range of poses, and the AFW dataset~\cite{zhu2012face}, with the comparison with a number of state-of-the-art methods.
We demonstrate superior $2$D alignment accuracy and quantitatively evaluate the $3$D alignment accuracy.

In summary, the main contributions of this work are:
\begin{itemize}
\item To the best of our knowledge, this is the first face alignment that can estimate $2$D/$3$D landmarks and their visibilities for a face image with an arbitrary pose.
\item By integrating with a $3$D deformable model, a cascaded coupled-regressor approach is developed to estimate both the camera projection matrix and the $3$D landmarks, where one benefit of $3$D model is the automatically computed landmark visibilities via surface normals.
\item A substantially larger number of non-frontal view face images are utilized in evaluation with demonstrated superior performances than state of the art.
\end{itemize}

\Section{Prior Work}

We now review the prior work in generic face alignment, pose-invariant face alignment, and $3$D face alignment.

The first type of face alignment approach is based on Constrained Local Model (CLM), where the first example is the ASM~\cite{Cootes1994}. 
The basic idea is to learn a set of local appearance models, one for each landmark, and the decision from the local models are combined with a global shape model. 
There are generative or discriminative~\cite{Cristinacce2007} approaches in learning the local model, and various approaches in utilizing the shape constraint~\cite{asthana2013robust}.
While the local models are favored for higher estimation precision, it also creates difficulty for alignment on low-resolution images due to limited local appearance.
In contrast, the AAM method~\cite{Cootes2001,Matthews2004} and its extension~\cite{lucey2013fourier,sanchez2012continuous} learn a global appearance model, whose similarity to the input image drives the estimation of the landmarks.
While the AAM is known to have difficulty with unseen subjects~\cite{Gross2005}, the recent development has substantially improved its generalization capability~\cite{tzimiropoulos2013optimization}.
Motivated by the Shape Regression Machine~\cite{Zhou2007} in the medical domain, cascaded regressor-based methods have been very popular in recent years~\cite{valstar2010facial,cao2014face}.
On one hand, the series of regressors progressively reduce the alignment error and lead to a high accuracy. 
On the other hand, advanced feature learning also renders ultra-efficient alignment procedures~\cite{ren2014face,kazemi2014one}. 
Other than the three major types of algorithms, there are also recent works based on graph-model~\cite{zhu2012face} and deep learning~\cite{zhang2014facial}.

\begin{figure*}[th!]
	\centering
	\includegraphics[width=0.9\textwidth]{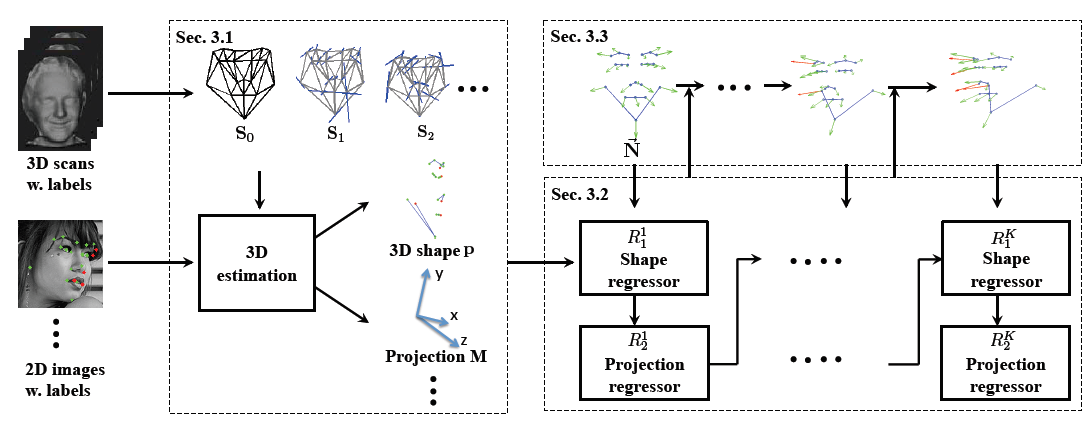}
	\caption{\small Overall architecture of our proposed PIFA method, with three main modules ($3$D modeling, cascaded coupled-regressor learning, and $3$D surface-enabled visibility estimation). Green/red arrows indicate surface normals pointing toward/away from the camera.}
	\figvspace
	\label{fig:overall}
\end{figure*}

Despite the explosion of methodology and efforts on face alignment, the literature on pose-invariant face alignment is rather limited, as shown in Tab.~\ref{tab:priorwork}.
There are four approaches explicitly handling faces with a wide range of poses.
Zhu and Ranaman proposed the TSPM approach for simultaneous face detection, pose estimation and face alignment~\cite{zhu2012face}.
An AFW dataset of in-the-wild faces with all poses is labeled with $6$ landmarks and used for experiments.
The cascaded deformable shape model (CDM) is a regression-based approach and probably the first approach claiming to be ``pose-free''~\cite{yu2013pose}, therefore it is the most relevant work to ours.
However, most of the experimental datasets contain near-frontal view images, except the AFW dataset with improved performance than~\cite{zhu2012face}. Also, there is no visibility estimation of the $2$D landmarks.
Zhang et al.~develop an effective deep learning based method to estimate $5$ landmarks.
While accurate results are obtained, all testing images appear to be within $\sim$$\pm60^{\circ}$ so that all $5$ landmarks are visible and there is no visibility estimation.
The OSRD approach has the similar experimental constraint in that all images are within $\pm40^{\circ}$~\cite{xing2014towards}.
Other than these four works, the work on occlusion-invariant face alignment are also relevant since non-frontal faces can be considered as one kind of occlusions, such as RCPR~\cite{burgos2013robust} and CoR~\cite{yu2014consensus}.
Despite being able to estimate visibilities, neither methods have been evaluated on faces with large pose variations. 
Finally, all aforementioned methods in this paragraph do not explicitly estimate the $3$D locations of landmarks.

$3$D face alignment aims to recover the $3$D location of facial landmarks given a $2$D image~\cite{Gu2006,wang2011viewpoint}.
There is also a very recently paper on $3$D face alignment from videos~\cite{Jeni15FG_ZFace}.
However, almost all methods take near-frontal-view face images as input, while our method can make use of all pose images.
A relevant but different problem is $3$D face reconstruction, which recovers the {\it detailed $3$D surface model} from one image, multiple images, and an image collection~\cite{Stylianou2006,gonzalez2010learning}. 
Finally, $3$D face model has been used in assisting $2$D face alignment~\cite{Xiao2004}.
However, it has not been explicitly integrated into the powerful cascaded regressor framework, which is the one of the main technical novelties of our approach.

\Section{Pose-Invariant 3D Face Alignment}


This section presents the details of our proposed Pose-Invariant $3$D Face Alignment (PIFA) algorithm, with emphasis on the training procedure.
As shown in Fig.~\ref{fig:overall}, we first learn a $3$D Morphable Model ($3$DMM) from a set of labeled $3$D scans, where a set of $2$D landmarks on an image can be considered as a projection of a $3$DMM instance (i.e., $3$D landmarks).
For each $2$D training face image, we assume that there exists the manual labeled $2$D landmarks and their visibilities, as well as the corresponding {\it $3$D ground truth}-- $3$D landmarks and the camera projection matrix.
Given the training images and $2$D/$3$D ground truth, we train a cascaded coupled-regressor that is composed of two regressors at each cascade layer, for the estimation of the update of the $3$DMM coefficient and the projection matrix respectively.
Finally, the visibilities of the projected $3$D landmarks are automatically computed via the domain knowledge of the $3$D surface normals, and incorporated into the regressor learning procedure.

\SubSection{3D Face Modeling}
\label{sec:3dFM}
Face alignment concerns the $2$D face shape, represented by the location of $N$ $2$D landmarks, i.e.,
\begin{equation}\label{eq:2dlandmarks}
\mathbf{U} = \left(
\begin{matrix}
  u_1 & u_2 &\cdots & u_N \\
  v_1 & v_2 & \cdots & v_N 
 \end{matrix}
\right),
\end{equation}
A $2$D face shape $\mathbf{U}$ is a projection of a $3$D face shape $\mathbf{S}$, similarly represented by the homogeneous coordinates of $N$ $3$D landmarks, i.e.,
\begin{equation}\label{eq:3dlandmarks}
\mathbf{S} = \left(
\begin{matrix}
  x_1 & x_2 &\cdots & x_N \\
  y_1 & y_2 & \cdots & y_N \\
  z_1 & z_2 & \cdots & z_N \\
  1 & 1 & \cdots & 1
 \end{matrix}
\right),
\end{equation}
 Similar to the prior work~\cite{Xiao2004}, a weak perspective model is assumed for the projection,
\begin{equation}\label{eq:project}
\mathbf{U} = \mathbf{M}\mathbf{S},
\end{equation}
where $\mathbf{M}$ is a $2\times4$ projection matrix with six degrees of freedom (yaw, pitch, row, scale and $2$D translation).

Following the basic idea of $3$DMM~\cite{Blanz2003}, we assume a $3$D face shape is an instance of the $3$DMM,
\begin{equation}\label{eq:3DASM}
\mathbf{S} = \mathbf{S}_0 + \sum_{i=1}^{N_s} p_i\mathbf{S}_i,
\end{equation}
where $\mathbf{S}_0$ and $\mathbf{S}_i$ is the mean shape and $i$th shape basis of the $3$DMM respectively, $N_s$ is the total shape bases, and $p_i$ is the $i$th shape coefficient.
Given a dataset of $3$D scans with manual labels on $N$ $3$D landmarks per scan, we first perform procrustes analysis on the $3$D scans to remove the global transformation, and then conduct Principal Component Analysis (PCA) to obtain the $\mathbf{S}_0$ and $\{\mathbf{S}_i\}$~\cite{Blanz2003} (see the top-left part of Fig.~\ref{fig:overall}).

The collection of all shape coefficients $\mathbf{p} = (p_1,p_2,\cdots,p_{N_s})$ is termed as the {\it $3$D shape parameter} of an image.
At this point, the face alignment for a testing image $\mathbf{I}$ has been converted from the estimation of $\mathbf{U}$ to the estimation of $\mathbf{P}=\{\mathbf{M},\mathbf{p}\}$.
The conversion is motivated by a few factors.
First, without the $3$D modeling, it is very difficult to model the out-of-plane rotation, which has a varying number of landmarks depending on the rotation angle.
Second, as pointed out by~\cite{Xiao2004}, by only using $\frac{1}{6}$ of the number of the shape bases, $3$DMM can have an equivalent representation power as its $2$D counterpart. 
Hence, using $3$D model might lead to a more compact representation of unknown parameters.

\Paragraph{Ground truth $\mathbf{P}$}
Estimating $\mathbf{P}$ for a testing image implies the existence of ground truth $\mathbf{P}$ for each training image.
However, while $\mathbf{U}$ can be manually labeled on a face image, $\mathbf{P}$ is normally unavailable unless a $3$D scan is captured along with a face image.
Therefore, in order to leverage the vast amount of existing $2$D face alignment datasets, such as the AFLW dataset~\cite{koestinger11b}, it is desirable to estimate $\mathbf{P}$ for a face image and use it as the ground truth for learning.

Given a face image $\mathbf{I}$, we denote the manually labeled $2$D landmarks as $\mathbf{U}$ and the landmark visibility as $\mathbf{v}$, an $N$-dim vector with binary elements indicating visible ($1$) or invisible ($0$) landmarks.
Note that for invisible landmarks, it is not necessary to label their $2$D location.
We define the following objective function to estimate $\mathbf{M}$ and $\mathbf{p}$,
\begin{equation}\label{eq:PEst}
J(\mathbf{M},\mathbf{p}) = \left|\left| \left(\mathbf{M}\left ( \mathbf{S}_0 + \sum_{i=1}^{N_s} p_i\mathbf{S}_i \right) - \mathbf{U} \right)\odot\mathbf{V} \right|\right|^2,
\end{equation}
where $\mathbf{V}=(\mathbf{v}^\T;\mathbf{v}^\T)$ is a $2\times N$ visibility matrix, $\odot$ denotes the element-wise multiplication, and $||\cdot||^2$ is the sum of the squares of all matrix elements.
Basically $J(\cdot,\cdot)$ computes the difference between the visible $2$D landmarks and their $3$D projections.
An alternative estimation scheme is utilized, i.e., by assuming $\mathbf{p}^0=0$, we estimate $\mathbf{M}^{k} = \argmin_{\mathbf{M}} J(\mathbf{M},\mathbf{p}^{k-1})$, and then $\mathbf{p}^{k} = \argmin_{\mathbf{p}} J(\mathbf{M}^k,\mathbf{p})$ iteratively until the changes on $\mathbf{M}$ and $\mathbf{p}$ are small enough.
Both minimizations can be efficiently solved in closed forms via least-square error.

\SubSection{Cascaded Coupled-Regressor}

For each training image $\mathbf{I}_i$, we now have its ground truth as $\mathbf{P}_i=\{\mathbf{M}_i,\mathbf{p}_i\}$, as well as their initialization, i.e., $\mathbf{M}_i^0=g(\mathbf{\bar{M}},\mathbf{b}_i)$ and $\mathbf{p}_i^0=\mathbf{0}$.
Here $\mathbf{\bar{M}}$ is the average of ground truth projection matrices in the training set, $\mathbf{b}_i$ is a $4$-dim vector indicating the bounding box location, and $g(\mathbf{M},\mathbf{b})$ is a function that modifies the scale and translation of $\mathbf{M}$ based on $\mathbf{b}$.
Given a dataset of $N_d$ training images, the question is {\it how} to formulate an optimization problem to estimate $\mathbf{P}_i$.
We decide to extend the successful cascaded regressors framework due to its accuracy and efficiency~\cite{cao2014face}. 
The general idea of cascaded regressors is to learn a series of regressors, where the $k$th regressor estimates the difference between the current parameter $\mathbf{P}_i^{k-1}$ and the ground truth $\mathbf{P}_i$, such that the estimated parameter gradually approximates the ground truth.

Motivated by this general idea, we adopt a cascaded coupled-regressor scheme where two regressors are learned at the $k$th cascade layer, for the estimation of $\mathbf{M}_i$ and $\mathbf{p}_i$ respectively.
Specifically, the first learning task of the $k$th regressor is,
\begin{equation}\label{eq:r1}
\Theta_1^k = \argmin_{\Theta_1^k} \sum_{i=1}^{N_d} ||\Delta\mathbf{M}_i^k - R_1^k( \mathbf{I}_i,  \mathbf{U}_i,  \mathbf{v}_i^{k-1}; \Theta_1^k ) ||^2,
\end{equation}
where 
\begin{equation}
\mathbf{U}_i = \mathbf{M}_i^{k-1}\left (\mathbf{S}_0 + \sum_{i=1}^{N_s} p_i^{k-1}\mathbf{S}_i \right),
\label{eq:ui}
\end{equation}
is the current estimated $2$D landmarks, $\Delta\mathbf{M}_i^k = \mathbf{M}_i - \mathbf{M}_i^{k-1}$, and  $R_1^k(\cdot;\Theta_1^k)$ is the desired regressor with the parameter of $\Theta_1^k$.
After $\Theta_1^k$ is estimated, we obtain $\Delta\mathbf{\hat{M}}_i = R_1^k(\cdot;\Theta_1^k)$ for all training images and update
$\mathbf{M}_i^k=\mathbf{M}_i^{k-1}+\Delta\mathbf{\hat{M}}_i$.
Note that this liner updating may potentially break the constraint of the projection matrix. 
Therefore, we estimate the scale and yaw, pitch, row angles ($s,\alpha,\beta,\gamma$) from $\mathbf{M}_i^k$ and compose a new $\mathbf{M}_i^k$ based on these four parameters.

Similarly the second learning task of the $k$th regressor is,
\begin{equation}\label{eq:r2}
\Theta_2^k = \argmin_{\Theta_2^k} \sum_{i=1}^{N_d} ||\Delta\mathbf{p}_i^k - R_2^k( \mathbf{I}_i,  \mathbf{U}_i,  \mathbf{v}_i^{k}; \Theta_2^k ) ||^2,
\end{equation}
where $\mathbf{U}_i$ is computed via Eq~\ref{eq:ui} except $\mathbf{M}_i^{k-1}$ is replaced with $\mathbf{M}_i^{k}$.
We also obtain $\Delta\mathbf{\hat{p}}_i = R_2^k(\cdot;\Theta_2^k)$ to all training images and update
$\mathbf{p}_i^k=\mathbf{p}_i^{k-1}+\Delta\mathbf{\hat{p}}_i$.
This iterative learning procedure continues for $K$ cascade layers. 

\Paragraph{Learning $R^k(\cdot)$}
Our cascaded coupled-regressor scheme does not depend on the particular feature representation or the type of regressors. 
Therefore, we may define them based on the prior work or any future development in features and regressors.
Specifically, in this work we adopt the HOG-based linear regressor~\cite{yan2013learn} and the fern regressor~\cite{burgos2013robust}. 

For the linear regressor, we denote a function $f(\mathbf{I}, \mathbf{U})$ to extract HOG features around a small rectangular region of each one of $N$ landmarks, which returns a $32N$-dim feature vector.
Thus, we define the regressor function as 
\begin{equation}\label{eq:r1linear}
R(\cdot) = \Theta^\T \cdot\mbox{Diag}^*(\mathbf{v}_i) f(\mathbf{I}_i, \mathbf{U}_i),
\end{equation}
where $\mbox{Diag}^*(\mathbf{v})$ is a function that duplicates each element of $\mathbf{v}$ $32$ times and converts into a diagonal matrix of size $32N$.
Note that we also add a constraint, $\lambda||\Theta||^2$, to Eq~\ref{eq:r1} or Eq~\ref{eq:r2} for a more robust least-square solution. 
By plugging Eq~\ref{eq:r1linear} to Eq~\ref{eq:r1} or Eq~\ref{eq:r2}, the regressor parameter $\Theta$ (e.g., a $N_s\times32N$ matrix for $R_2^k$) can be easily estimated in the closed form.  

For the fern regressor, we follow the training procedure of~\cite{burgos2013robust}. 
That is, we divide the face region into a $3\times3$ grid. 
At each cascade layer, we choose $3$ out of $9$ zones with the least occlusion, computed based on the $\{\mathbf{v}_i^k\}$. 
For each selected zone, a depth $5$ random fern regressor is learned from the shape-index features selected by correlation-based method~\cite{cao2014face} from that zone only. 
Finally the learned $R(\cdot)$ is a weighted mean voting from the $3$ fern regressors, where the weight is inversely proportional to the average amount of occlusion in that zone.


\SubSection{3D Surface-Enabled Visibility}

Up to now the only thing that has not been explained in the training procedure is the visibility of the projected $2$D landmarks, $\mathbf{v}_i$.
It is obvious that during the testing we have to estimate $\mathbf{v}$ at each cascade layer for each testing image, since there is no visibility information given.
As a result, during the training procedure, we also have to {\it estimate} $\mathbf{v}$ per cascade layer for each {\it training image}, rather than using the manually labeled ground truth visibility that is useful for estimating ground truth $\mathbf{P}$ as shown in Eq~\ref{eq:PEst}.

Depending on the camera projection matrix $\mathbf{M}$, the visibility of each projected $2$D landmark can dynamically change along different layers of the cascade (see the top-right part of Fig.~\ref{fig:overall}).
In order to estimate $\mathbf{v}$, we decide to use the $3$D face surface information.
We start by assuming every individual has a similar $3$D surface normal vector at each of its $3$D landmarks.
Then, by rotating the surface normal according to the rotation angle indicated by the projection matrix, we can know that whether the coordinate of the $z$-axis is pointing toward the camera (i.e., visible) or away from the camera (i.e., invisible). 
In other words, the sign of the $z$-axis coordinates indicates visibility.

By taking a set of $3$D scans with manually labeled $3$D landmarks, we can compute the landmarks' average $3$D surface normals, denoted as a $3\times N$ matrix $\vec{\mathbf{N}}$.
Then we use the following equation to compute the visibility vector,
\begin{equation}\label{eq:v}
\mathbf{v} =  \vec{\mathbf{N}}^\T\cdot \left( \frac{\mathbf{m}_1}{ ||\mathbf{m}_1|| } \times \frac{\mathbf{m}_2}{ ||\mathbf{m}_2|| } \right),
\end{equation}
where $\mathbf{m}_1$ and $\mathbf{m}_2$ are the left-most three elements at the first and second row of $\mathbf{M}$ respectively, and $||\cdot||$ denotes the $L_2$ norm.
For fern regressors, $\mathbf{v}$ is a soft visibility within $\pm1$.
For linear regressors, we further compute $\mathbf{v} = \frac{1}{2}(1+\mbox{sign}(\mathbf{v}))$, which results in a hard visibility of either $1$ or $0$.

In summary, we present the detailed training procedure in Algorithm~\ref{alg:training}.

\begin{algorithm}[t!]
\small
\KwData{$3$D model $\{\{\mathbf{S}\}_{i=0}^{N_s}, \vec{\mathbf{N}} \}$, training samples and labels $\{\mathbf{I}_i, \mathbf{U}_i, \mathbf{b}_i\}_{i=1}^{N}$.}
\KwResult{Cascaded coupled-regressor parameters $\{\Theta_1^k, \Theta_2^k \}_{k=1}^K$.}


\ForEach{$i = 1, \cdots, N_d$}{
	Estimate $\mathbf{M}_i$ and $\mathbf{p}_i$ via Eq.~\ref{eq:PEst}\;

	$\mathbf{M}_i^0=g(\mathbf{\bar{M}},\mathbf{b}_i)$, $\mathbf{p}_i^0 = \mathbf{0}$ and $\mathbf{v}_i^0 = \mathbf{1}$ \;

}

\ForEach{$k = 1, \cdots, K$}{

	Compute $\mathbf{U}_i$ via Eq~\ref{eq:ui} for each image \;

	Estimate $\Theta_1^k$ via Eq~\ref{eq:r1} \;

	Update $\mathbf{M}_i^k$ and $\mathbf{U}_i$ for each image \;

	Compute $\mathbf{v}_i^{k}$ via Eq~\ref{eq:v} for each image \;

	Estimate $\Theta_2^k$ via Eq~\ref{eq:r2} \;

	Update $\mathbf{p}_i^k$ for each image \;

}

\Return{$\{R_1^k(\cdot;\Theta_1^k), R_2^k(\cdot;\Theta_2^k) \}_{k=1}^K$.}

\caption{The training procedure of PIFA.}
\label{alg:training}
\end{algorithm}

\Paragraph{Model Fitting}
Given a testing image $\mathbf{I}$ and its initial parameter $\mathbf{M}^0$ and $\mathbf{p}^0$, we can apply the learned cascaded coupled-regressor for face alignment.
Basically we iteratively use $R_1^k(\cdot;\Theta_1^k)$ to compute $\Delta{\hat{\mathbf{M}}}$, update $\mathbf{M}^k$, use $R_2^k(\cdot;\Theta_2^k)$ to compute $\Delta{\hat{\mathbf{p}}}$, and update $\mathbf{p}^k$. 
Finally the estimated $3$D landmarks are $\mathbf{\hat{S}} = \mathbf{S}_0+\sum_i p_i^K \mathbf{S}_i$, and the estimated $2$D landmarks are $\mathbf{\hat{U}} = \mathbf{M}^K \mathbf{\hat{S}}$.
Note that $\mathbf{\hat{S}}$ carries the individual $3$D shape information of the subject, but not necessary in the same pose as the $2$D testing image.

\Section{Experimental Results}

\Paragraph{Datasets}
The goal of this work is to advance the capability of face alignment on {\bf in-the-wild faces with all possible view angles}, which is the type of images we desire when selecting experimental datasets. 
However, very few publicly available datasets satisfy this characteristic, or have been extensively evaluated in prior work (see Tab.~\ref{tab:priorwork}).
Nevertheless, we identify three datasets for our experiments. 

AFLW dataset~\cite{koestinger11b} contains $\sim$$25k$ in-the-wild face images, each image annotated with the {\it visible} landmarks (up to $21$ landmarks), and a bounding box. 
Based on our estimated $\mathbf{M}$ for each image, we select a subset of $5,300$ images where the numbers of images whose absolute yaw angle within $[0^{\circ},30^{\circ}]$, $[30^{\circ},60^{\circ}]$, $[60^{\circ},90^{\circ}]$ are roughly $\frac{1}{3}$ each.
To have a more {\it balanced distribution} of the left vs.~right view faces, we take the odd indexed images among $5,300$ (i.e., $1$st, $3$rd,...), flip the images horizontally, and use them to replace the original images.
Finally, a random partition leads to $3,901$ and $1,299$ images for training and testing respectively.
Note that from Tab.~\ref{tab:priorwork}, it is clear that among all methods that test on all poses, we have the largest number of testing images.

AFW dataset~\cite{zhu2012face} contains $205$ images and in total $468$ faces with different poses within $\pm90^{\circ}$. 
Each image is labeled with {\it visible} landmarks (up to $6$), and a face bounding box. 
We only use AFW for testing.

Since we are also estimating $3$D landmarks, it is important to test on a dataset with {\it ground truth}, rather than estimated, $3$D landmark locations.
We find BP4D-S database~\cite{BP4D-Spontanous} to be the best for this purpose, which contains pairs of $2$D images and $3$D scans of spontaneous facial expressions from $41$ subjects. 
Each pair has semi-automatically generated $83$ $2$D and $83$ $3$D landmarks, and the pose.
We apply a random perturbation on $2$D landmarks (to mimic imprecise face detection) and generate their enclosed bounding box.
With the goal of selecting as many non-frontal view faces as possible, we select a subset where the numbers of faces whose yaw angle within $[0^{\circ},10^{\circ}]$, $[10^{\circ},20^{\circ}]$, $[20^{\circ},30^{\circ}]$ are $100$, $500$, and $500$ respectively.
We randomly select half of $1,100$ images for training and the rest for testing, with disjoint subjects.


\Paragraph{Experiment setup}
Our PIFA approach needs a $3$D model of $\{\mathbf{S}\}_{i=0}^{N_s}$ and $\vec{\mathbf{N}}$.
Using the BU-4DFE database~\cite{BU-4DFE} that contains $606$ $3$D facial expression sequences from $101$ subjects, we evenly sample $72$ scans from each sequence and gather a total of $72\times606$ scans.  
Based on the method in Sec.~\ref{sec:3dFM}, the resultant model has $N_s=30$ for AFLW and AFW, and $N_s=200$ for BP4D-S. 

During the training and testing, for each image with a bounding box, we place the mean $2$D landmarks (learned from a specific training set) on the image such that the landmarks on the boundary overlap with the four edges of the box.
For training with linear regressors, we set $K=10$, $\lambda=120$, while $K=150$ for fern regressors.

\Paragraph{Evaluation metric}
Given the ground truth $2$D landmarks $\mathbf{U}_i$, their visibility $\mathbf{v}_i$, and estimated landmarks $\mathbf{\hat{U}}_i$ of $N_t$ testing images, we have two ways of computing the landmark estimation errors:
1) Mean Average Pixel Error (MAPE)~\cite{yu2013pose}, which is the average of the estimation errors for visible landmarks, i.e., 
\begin{equation}\label{eq:mape}
\mbox{MAPE} = \frac{1}{\sum_i^{N_t} |\mathbf{v}_i|_1 }\sum_{i,j}^{N_t,N}\mathbf{v}_i(j)||\mathbf{\hat{U}}_i(:,j) - \mathbf{U}_i(:,j)||,
\end{equation}
where $|\mathbf{v}_i|_1$ is the number of visible landmarks of image $\mathbf{I}_i$, and $\mathbf{U}_i(:,j)$ is the $j$th column of $\mathbf{U}_i$.
2) Normalized Mean Error (NME), which is the average of the normalized estimation error of visible landmarks, i.e.,
\begin{equation}\label{eq:nme}
\mbox{NME} = \frac{1}{N_t}\sum_{i}^{N_t}( \frac{1}{d_i |\mathbf{v}_i|_1 }\sum_j^N \mathbf{v}_i(j)||\mathbf{\hat{U}}_i(:,j) - \mathbf{U}_i(:,j)||).
 \end{equation}
where $d_i$ is the square root of the face bounding box size, as used by~\cite{yu2013pose}.
Note that normally $d_i$ is the distance of two centers of eyes in most prior face alignment work dealing with near-frontal face images.

Given the ground truth $3$D landmarks $\mathbf{S}_i$ and estimated landmarks $\mathbf{\hat{S}}_i$, we first estimate the global rotation, translation and scale transformation so that the transformed $\mathbf{S}_i$, denoted as $\mathbf{S}'_i$, has the minimum distance to $\mathbf{\hat{S}}_i$. 
We then compute the MAPE via Eq~\ref{eq:mape} except replacing $\mathbf{U}$ and $\mathbf{\hat{U}}_i$ with $\mathbf{S}'_i$ and $\mathbf{\hat{S}}_i$, and $\mathbf{v}_i=\mathbf{1}$.
Thus the MAPE only measures the error due to non-rigid shape estimation, rather than the pose estimation.

\Paragraph{Choice of baseline methods}
Given the explosion of face alignment work in recent years, it is important to choose appropriate baseline methods so as to make sure the proposed method advances the state of the art.
In this work, we select three recent works as baseline methods:
1) CDM~\cite{yu2013pose} is a CLM-type method and the first one claimed to perform pose-free face alignment, which has exactly the same objective as ours. 
On AFW it also outperforms the other well-known TSPM method~\cite{zhu2012face} that can handle all pose faces.
2) TCDCN~\cite{zhang2014facial} is a powerful deep learning-based method published in the most recent ECCV.
Although it only estimates $5$ landmarks for up to $\sim$$60^{\circ}$ yaw, it represents the most recent development in face alignment.
3) RCPR~\cite{burgos2013robust} is a regression-type method that represents the occlusion-invariant face alignment. 
Although it is an earlier work than CoR~\cite{yu2014consensus}, we choose it due to its superior performance on the large COFW dataset (see Tab.~1 of~\cite{yu2014consensus}).
It can be seen that these three baselines not only are most relevant to our focus on pose-invariant face alignment, but also well represent the major categories of existing face alignment algorithms based on~\cite{wang2014facial}.

\Paragraph{Comparison on AFLW}
\begin{table}[t!]
	\centering
	\small
	\caption{\small The NME(\%) of three methods on AFLW.}
	\begin{tabular}{c|c|c|c}
		\hline						

	 $N_t$  &      PIFA       &  CDM    & RCPR  \\ \hline
$1,299$ &  $\mathbf{6.52}$ &   &  $7.15$ \\ \hline
$783$&  $\mathbf{6.08}$ & $8.65$ &  \\ \hline							 
	       \hline
	\end{tabular}
	\label{tab:alfw}
\end{table}
Since the source code of RCPR is publicly available, we are able to perform the training and testing of RCPR on our specific AFLW partition. We use the available executable of CDM to compute its performance on our test set .
We strive to provide the same setup to the baselines as ours, such as the initial bounding box, regressor learning, etc. 
For our PIFA method, we use the fern regressor. 
Because CDM integrates face detection and pose-free face alignment, no bounding box was given to CDM and it successfully detects and aligns $783$ out of $1,299$ testing images. 
Therefore, to compare with CDM, we evaluate the NME on the {\it same} $783$ testing images.
As shown in Tab.~\ref{tab:alfw}, our PIFA shows superior performance to both baselines.
Although TCDCN also reports performance on a subset of $3,000$ AFLW images within $\pm60^{\circ}$ yaw, it is evaluated with $5$ landmarks, based on NME when $d_i$ is the inter-eye distance.
Hence, without the source code of TCDCN, it is difficult to have a fair comparison on our subset of AFLW images (e.g., we can not define $d_i$ as the inter-eye distance due to profile view images).
On the $1,299$ testing images, we also test our method with linear regressors, and achieve a NME of $7.50$, which shows the strength of fern regressors.

\Paragraph{Comparison on AFW}
\begin{table}[t!]
	\centering
	\small
	\caption{\small The comparison of four methods on AFW.}
	\resizebox{\linewidth}{!} {
	\begin{tabular}{c|c|c|c|c|c|c}
		\hline						

	 $N_t$  &  $N$ & Metric &     PIFA      &  CDM    & RCPR & TCDCN \\ \hline
$468$ & $6$  & MAPE & $\mathbf{8.61}$ &  $9.13$ &  &  \\ \hline
$313$ & $5$  & NME &  $\mathbf{7.9}$ & & $9.3$ & $8.2$  \\ \hline							
	       \hline
	\end{tabular}
	}
	\label{tab:afw}
\end{table}
Unlike our specific subset of AFLW, the AFW dataset has been evaluated by all three baselines, but different metrics are used.
Therefore, the results of the baselines in Tab.~\ref{tab:afw} are from the published papers, instead of executing the testing code.
One note is that from the TCDCN paper~\cite{zhang2014facial}, it appears that all $5$ landmarks are visible on all displayed images and no visibility estimation is shown, which
might suggest that TCDCN was evaluated on a subset of AFW with up to $\pm60^{\circ}$ yaw.
Hence, we select the total of $313$ out of $468$ faces within this pose range and test our algorithm.
Since it is likely that our subset could differ to~\cite{zhang2014facial}, please take this into consideration while comparing with TCDCN.
Overall, our PIFA method still performs favorably among the four methods.
This is especially encouraging given the fact that TCDCN utilizes a substantially larger training set of $10,000$ images - more than two times of our training set.
Note that in addition to Tab.~\ref{tab:alfw} and~\ref{tab:afw}, our PIFA also has other benefits as shown in Tab.~\ref{tab:priorwork}.
E.g., we have $3$D and visibility estimation, while RCPR has no $3$D estimation and TCDCN does not have visibility estimation.  

\begin{figure}[t!]
	\centering
	\includegraphics[width=0.9\linewidth]{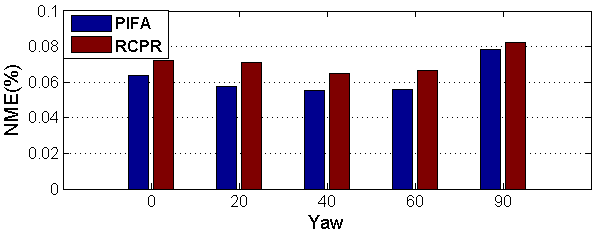}
	\caption{\small The NME of five pose groups for two methods.}
	\figvspace
	\label{fig:pose}
\end{figure}

\Paragraph{Estimation error across poses}
Just like pose-invariant face recognition studies the recognition rate across poses~\cite{chai2007locally}, we also like to study the performance of face alignment across poses.
As shown in Fig.~\ref{fig:pose}, based on the estimated projection matrix $M$ and its yaw angles, we partition all testing images of AFLW into five bins, each around a specific yaw angle.
Then we compute the NME of testing images within each bin, for our method and RCPR.
We can observe that the profile view images have in general larger NME than near-frontal images, which shows the challenge of pose-invariant face alignment. 
Further, the improvement of PIFA over RCPR is consistent across most of the poses.

\begin{figure}[t!]
	\centering
	\includegraphics[width=0.9\linewidth]{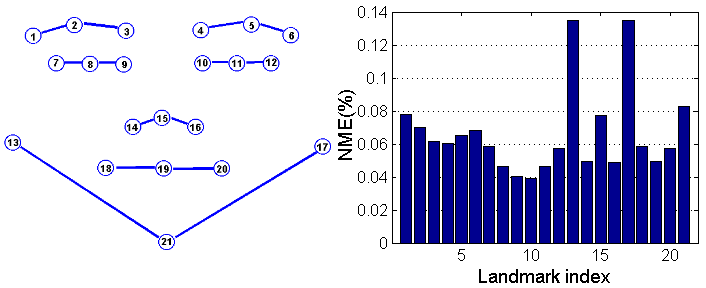}
	\caption{\small The NME of each landmark for PIFA.}
	\figvspace
	\label{fig:landmark}
\end{figure}

\Paragraph{Estimation error across landmarks}
We are also interested in the estimation error across various landmarks, under a wide range of poses.
Hence, for the AFLW test set, we compute the NME of each landmark for our method.
As shown in Fig.~\ref{fig:landmark}, the two eye region has the least amount of error.
The two landmarks under the ears have the most error, which is consistent with our intuition.
These observations also align well with prior face alignment study on near-frontal images.

\Paragraph{3D landmark estimation}
\begin{figure}[t!]
	\centering
	\includegraphics[width=0.9\linewidth]{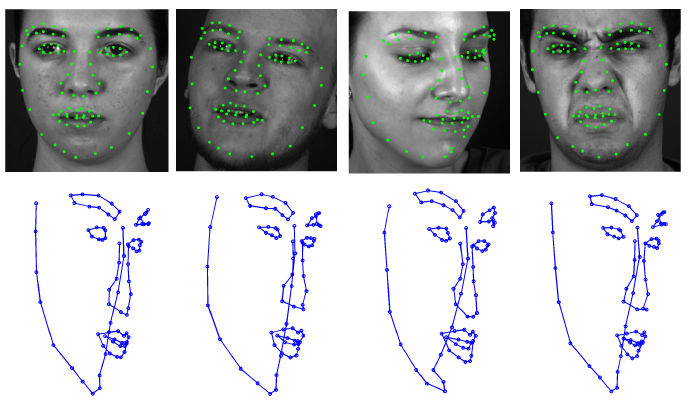}
	\caption{\small $2$D and $3$D alignment results of the BP4D-S dataset.}
	\label{fig:3d}
	\figvspace
\end{figure}
\begin{figure*}[t!]
	\centering
	\begin{tabular}{c}
	\includegraphics[ width=0.98\textwidth]{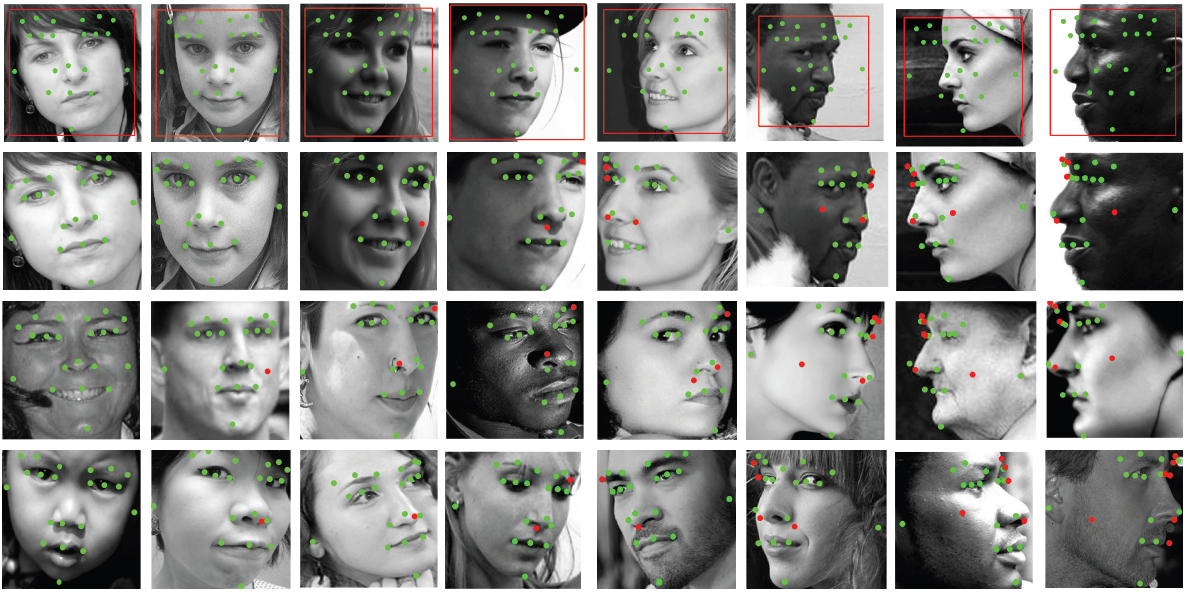} \\ \hline
	\includegraphics[width=0.98\textwidth]{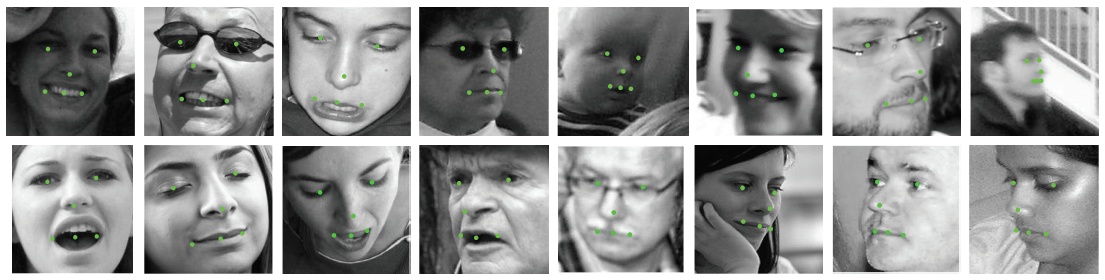} 
	\end{tabular}
	\vspace{-4mm}
	\caption{\small Testing results of AFLW (top) and AFW (bottom). As shown in the top row, we initialize face alignment by placing a $2$D mean shape in the given bounding box of each image. Note the {\it disparity} between the initial landmarks and the final estimated ones, as well as the diversity in pose, illumination and resolution among the images. Green/red points indicate visible/invisible estimated landmarks. }
		\figvspace

	\label{fig:example}
\end{figure*}
By performing the training and testing on the BP4D-S dataset, we can evaluate the MAPE of $3$D landmark estimation, with exemplar results shown in Fig.~\ref{fig:3d}.
Since there are limited $3$D alignment work and many of which do not perform quantitative evaluation, such as~\cite{Gu2006}, we are not able to find another method as the baseline. 
Instead, we use the $3$D mean shape, $\mathbf{S}_0$ as a baseline and compute its MAPE with respect to the ground truth $3$D landmarks $\mathbf{S}_i$ (after global transformation).
We find that the MAPE of $\mathbf{S}_0$ baseline is $5.02$, while our method has $4.75$.
Although our method offers a better estimation than the mean shape, this shows that $3$D face alignment is still a very challenging problem.
And we hope our efforts to quantitatively measure the $3$D estimation error, which is more difficult than its $2$D counterpart, will motivate more research activities to address this challenge.

\Paragraph{Computational efficiency}
\begin{table}[t!]
	\centering
	\small
	\caption{\small Efficiency of four methods in FPS.}
	\begin{tabular}{c|c|c|c}
		\hline						

	   PIFA      &  CDM    & RCPR & TCDCN \\ \hline
$3.0$ & $0.2$ &  $3.0$ & $\mathbf{58.8}$  \\ \hline		 
	       \hline
	\end{tabular}
	\label{tab:speed}
\end{table}
Based on the efficiency reported in the publications of baseline methods, we compare the computational efficiency of four methods in Tab.~\ref{tab:speed}.
Only TCDCN is measured based on the C implementation while other three are all based on Matlab implementation.
It can be observed that TCDCN is the most efficient one. 
Our unoptimized implementation has a reasonable speed of $3$ FPS and we believe this efficiency can be substantially improved with optimized C implementation.

\Paragraph{Qualitative results}
We now show the qualitative face alignment results for images in three datasets.
As shown in Fig.~\ref{fig:example}, despite the large pose range of $\pm90^{\circ}$ yaw, our algorithm does a good job of aligning the landmarks, and correctly predict the landmark visibilities. 
These results are especially impressive if you consider the same mean shape ($2$D landmarks) is used as the initilization of all testing images, which has very large deformations with respect to their final estimation landmarks.

\Section{Conclusions}

Motivated by the fast progress of face alignment technologies and the need to align faces at all poses, this paper draws attention to a relatively unexploited problem of face alignment robust to poses variation.
To this end, we propose a novel approach to tightly integrate the powerful cascaded regressor scheme and the $3$D deformable model.
The $3$DMM not only serves as a compact constraint, but also offers an automatic and convenient way to estimate the visibilities of $2$D landmarks - a key for successful pose-invariant face alignment.
As a result, for a $2$D image, our approach estimates the locations of $2$D and $3$D landmarks, as well as their $2$D visibilities.
We conduct an extensive experiment on a large collection of all-pose face images and compare with three state-of-the-art methods.
While superior $2$D landmark estimation has been shown, the performance on $3$D landmark estimation indicates the future direction to improve this line of work.

{\footnotesize
\bibliographystyle{ieee}
\bibliography{abbrev_brief,xl-literature}
}

\end{document}